\documentclass[10pt,twocolumn,letterpaper]{article}

\usepackage{iccv}
\usepackage{times}
\usepackage{epsfig}
\usepackage{graphicx}
\usepackage{amsmath}
\usepackage{amssymb}
\usepackage{dsfont}
\usepackage{subfigure}
\usepackage{mathrsfs}
\usepackage{bm}
\usepackage{color}
\usepackage{multirow}
\newcommand\equalcontribution{\thanks{Authors contributed equally and are listed in alphabetical order.}}

\usepackage[breaklinks=true,bookmarks=false,colorlinks=true]{hyperref}

\iccvfinalcopy 

\def\iccvPaperID{****} 

\ificcvfinal\fi

\begin{document}

\title{Pose-aware Multi-level Feature Network for Human Object \\Interaction Detection}

\author{ Bo Wan\equalcontribution \hspace{0.4cm} Desen Zhou\footnotemark[1] \hspace{0.3cm}  Yongfei Liu  \hspace{0.3cm} Rongjie Li \hspace{0.3cm} Xuming He \\
 ShanghaiTech University, Shanghai, China \\
	{\tt\small \{wanbo, zhouds, liuyf3, lirj2, hexm\}@shanghaitech.edu.cn}
}

\maketitle
\ificcvfinal\thispagestyle{empty}\fi
\begin{abstract}
	
	Reasoning human object interactions is a core problem in human-centric scene understanding and detecting such relations poses a unique challenge to vision systems due to large variations in human-object configurations, multiple co-occurring relation instances and subtle visual difference between relation categories.
	To address those challenges, we propose a multi-level relation detection strategy that utilizes human pose cues to capture global spatial configurations of relations and as an attention mechanism to dynamically zoom into relevant regions at human part level. Specifically, we develop a multi-branch deep network to learn a pose-augmented relation representation at three semantic levels, incorporating interaction context, object features and detailed semantic part cues.
	As a result, our approach is capable of generating robust predictions on fine-grained human object interactions with interpretable outputs. 
    Extensive experimental evaluations on public benchmarks show that our model outperforms prior methods by a considerable margin, demonstrating its efficacy in handling complex scenes.  Code is available at ~\href{https://github.com/bobwan1995/PMFNet}{https://github.com/bobwan1995/PMFNet}.
 \end{abstract}

\vspace{-5mm}
\section{Introduction}
Visual relations play an essential role in a deeper understanding of visual scenes, which usually requires reasoning beyond merely recognizing individual scene entities~\cite{sadeghi2011recognition, krishna2017visual, zhang2017visual}. Among different types of visual relations, human object interactions are ubiquitous in our visual environment and hence its inference is critical for many vision tasks, such as activity analysis~\cite{caba2015activitynet}, video understanding~\cite{ yu2016video} and visual question answering~\cite{balanced_vqa_v2}. 

The task of human object interaction (HOI) detection aims to localize and classify triplets of human, object and relation from an input image. 
While deep neural networks have led to significant progresses in object and action recognition~\cite{resnet,ren2015faster,feichtenhofer2016convolutional}, it remains challenging to detect HOIs due to large variations of human-object appearance and spatial configurations, multiple co-existing relations and subtle differences between similar relations~\cite{VCOCO,HOI-DET}.   

\begin{figure}[t]
	\centering
	\includegraphics[width=0.85\linewidth]{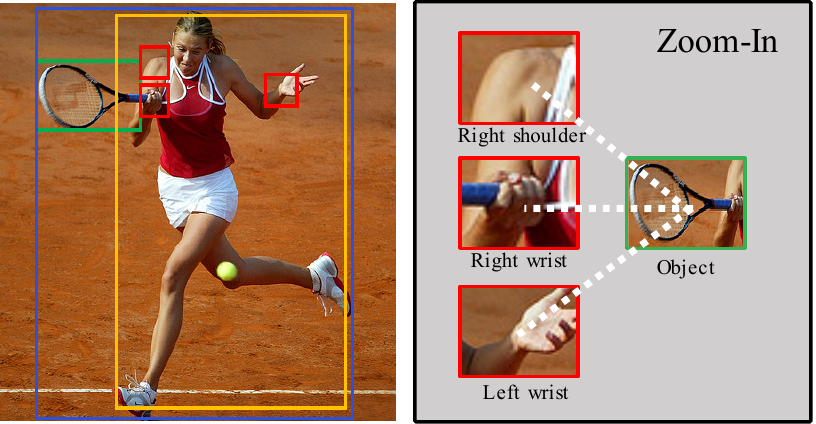}
	\caption{Our framework utilizes three levels of representation, including i): interaction ({\color{blue}{blue box}}), ii): visual object (\textcolor[RGB]{0,176,80}{green box} and \textcolor[RGB]{249,192,1}{yellow box}), iii): human parts (\textcolor[RGB]{244,1,4}{red boxes}) to recognize interaction. The highlight of our framework is \textit{human parts} level representation, which can provide discriminative feature. Here several informative human parts, like `right shoulder', `right wrist' and `left wrist', are focused to help recognize action `hold'.} \vspace{-5mm}
	\label{ads}
\end{figure}

Most of existing works on HOI detection tackle the problem by reasoning interactions at the visual object level~\cite{gkioxari2018detecting,gao2018ican,qi2018learning}. The dominant approaches typically start from a set of human-object proposals, and extract visual features of human and object instances which are combined with their spatial cues (e.g., masks of proposals) to predict relation classes of those human-object pairs~\cite{gao2018ican, xu2018interact,li2018transferable}. Despite their encouraging results, such coarse-level reasoning suffers from several drawbacks when handling relatively complex relations. 
First, it is difficult to determine the relatedness of a human-object pair instance with an object-level representation due to lack of context cues, which can lead to erroneous association. 
In addition, many relation types are defined in terms of fine-grained actions, which are unlikely to be differentiated based on similar object-level features. For instance, it may require a set of detailed local features to tell the difference between `\textit{hold}' and `\textit{catch}' in sport scenes. 
Furthermore, as these methods largely rely on holistic features, the reasoning process of relations is a blackbox and hard to interpret. 

In this work, we propose a new multi-level relation reasoning strategy to address the aforementioned limitations. Our main idea is to utilize estimated human pose to capture global spatial configuration of relations and as a guidance to extract local features at semantic part level for different HOIs. 
Such augmented representation enables us to incorporate interaction context, human-object and detailed semantic part cues into relation inference, and hence generate robust and fine-grained predictions with interpretable attentions.
To this end, we perform relation reasoning at three distinct semantic levels for each human-object proposal: i) interaction, ii) visual objects, and iii) human parts. Fig.~\ref{ads} illustrates an example of our relation reasoning. 

Specifically, at the \textit{interaction} level of a human-object proposal, we take a union region of the human and object instance, which encodes the context of the relation proposal, to produce an affinity score of the human-object pair. This score indicates how likely there exists a visual relation between the human-object pair and helps us to eliminate background proposals. For the \textit{visual object} level, we adopt a common object-level representation as in~\cite{HOI-DET, gao2018ican, li2018transferable} but augmented by human pose, to encode human-object appearance and their relative positions. 
The main focus of our design is a new representation at the \textit{human part} level, in which we use the estimated human pose to describe the detailed spatial and appearance cues of the human-object pair. To achieve this, we exploit the correlation between parts and relations to produce a part-level attention, which enable us to focus on sub-regions that are informative to each relation type. In addition, we compute the part locations relative to the object entity to encode a fine-level spatial configuration. Finally, we integrate the HOI cues from all three levels to predict the category of the human-object proposal.

We develop a multi-branch deep neural network to instantiate our multi-level relation reasoning, which consists of four main modules: a backbone module, a holistic module, a zoom-in module and a fusion module. Given an image, the \textit{backbone module} computes its convolution feature map, and generates human-object proposals and spatial configurations. For each proposal, the \textit{holistic module} integrates human, object and their union features, as well as an encoding of human pose and object location. The \textit{zoom-in module} extracts the human part and object features, and produces a part-level attention from the pose layout to enhance relevant part cues. The \textit{fusion module} combines the holistic and part-level representations to generate final scores for HOI categories.
We refer to our model as {\textit{Pose-aware Multi-level Feature Network} {\textbf(PMFNet)}}.
Given human-object proposals and pose estimation, our deep network is trained in an end-to-end fashion.  

We conduct extensive evaluations on two public benchmarks V-COCO and HICO-DET, and outperform the current state-of-the-art by a sizable margin. To better understand our method, we also provide detailed ablative study of our deep network on the V-COCO dataset. 

Our main contributions are three-folds:
\begin{itemize}
	\setlength{\itemsep}{0pt}
	\setlength{\parskip}{0pt}
	\setlength{\parsep}{0pt}  
	\item We propose a multi-level relation reasoning for human object interaction detection, in which we utilize human pose to capture global configuration and as an attention to extract detailed local appearance cues.
	\item We develop a modularized network architecture for HOI prediction, which produces an interpretable output based on relation affinity and part attention. 
	\item Our approach achieves the state-of-the-art performance on both V-COCO and HICO-DET benchmarks. 
\end{itemize}

\vspace{-2mm}
\section{Related Work}
\textbf{Visual Relationship Detection.} Visual Relation Detection (VRD)~\cite{lu2016visual, sadeghi2011recognition, krishna2017visual, zhang2017visual} aims at detecting the objects and describing their interactions simultaneously for a given image, which is a critical task for achieving visual scene understanding. Lu \emph{et al.}~\cite{lu2016visual} propose to learn language priors from semantic word embedding to fine-tune visual relationships. Zhang \emph{et al.}~\cite{zhang2017visual} design a visual translation network to embed objects into low-dimension relation space for tackling visual relationship detection problem. Besides, Xu~\emph{et al.}~\cite{xu2017scenegraph} model the visual relationship detection in structured scene as a graph, and propagate messages between objects. In our task, we focus on human-centric relationship detection, which aims to detect human-object interactions.

\begin{figure*}[th!]
    \centering
     \includegraphics[width=0.8\linewidth]{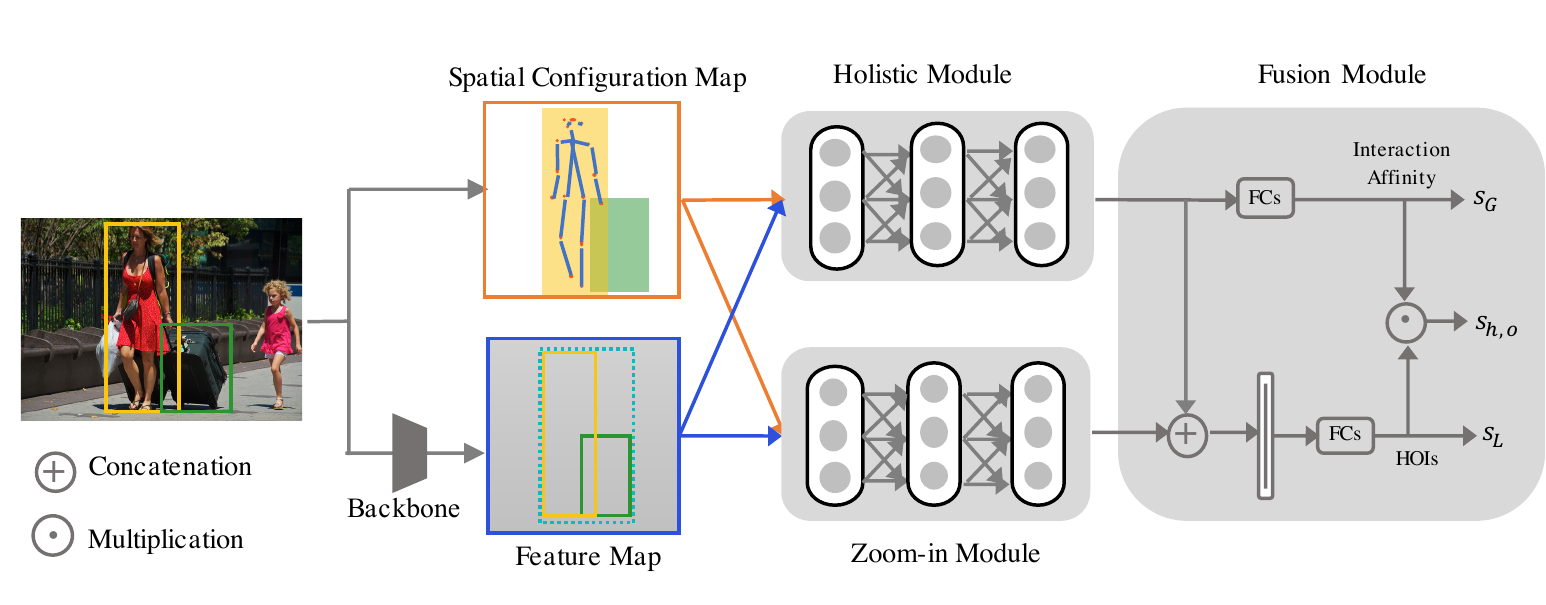}
    \caption[]{Overview of our framework: For a pair of human-object proposals and related human pose, \textbf{Backbone Module} aims to prepare convolution feature map and Spatial Configuration Map~(SCM). \textbf{Holistic Module}  generates object-level features and \textbf{Zoom-in Module} captures part-level features. Finally \textbf{Fusion Module} combines object-level and part-level cues to predict final scores for HOI categories.  }\vspace{-4mm}
    \label{fig:model} 
\end{figure*}

\textbf{Human-Object Interaction Detection.} Human-Object interaction (HOI) detection is essential for understanding human behaviors in a complex scene. In recent years, researchers have developed several human object interaction dataset, like V-COCO~\cite{VCOCO} and HICO-DET~\cite{HOI-DET}. Early studies mainly focus on tackling HOIs recognition by utilizing multi-stream information, including human, object appearance, spatial information and human poses. In HO-RCNN~\cite{HOI-DET}, Chao \emph{et al.} propose multi-stream to aggregate human, object and spatial configuration information to resolve HOIs detection tasks. Qi \emph{et al.}~\cite{qi2018learning} propose graph parsing neural network (GPNN) to model the structured scene into a graph and propagate messages between each human and object node and classify all nodes and edges for their possible object classes and actions. 

There have been several attempts that use human pose for recognizing fine-grained human related actions~\cite{fang2018pairwise,gao2018ican, li2018transferable}. Fang \emph{et al.}~\cite{fang2018pairwise} exploit pair-wise human parts correlation to help to tackle HOIs detection. Li \emph{et al.}~\cite{li2018transferable} explore interactiveness prior existing in multiple datasets and combine human pose and spatial configuration to form pose configuration map. However, those works only take human pose as a spatial constraint between human parts and object, but do not use it to extract zoom-in feature in each part, which provides more detail information for HOI task. By contrast, we take advantage of this fine-grained feature to capture subtle differences between similar interactions.

\textbf{Attention Models.} Attention mechanism has been proved very effective in various vision tasks, including image captioning~\cite{xu2015show, you2016image}, fine grained classification~\cite{ji2018stacked}, pose estimation~\cite{chu2017multi} and action recognition\cite{sharma2015action, girdhar2017attentional}. Attention mechanism can help highlight informative regions or parts and suppress some irrelevant global information. Xu \emph{et al.}~\cite{xu2015show} firstly utilize attention mechanism in image captioning and automatically attend some informative region in image relevant to generated sentences. Sharma \emph{et al.}~\cite{sharma2015action} apply attention models  realized by LSTM into action recognition task to learn important parts of video frames. Yu \emph{et al.}~\cite{ji2018stacked} propose a stacked semantic guided attention to focus on informative birds' parts and suppress irrelevant global information. In our work, we focus on pose-aware attention for HOIs detection in human parts. 


\vspace{-2mm}

\section{Method}
We now introduce our multi-level relation reasoning strategy for human-object interaction detection. Our goal is to localize and recognize human-object interaction instances in an image. To this end, we augment object-level cues with human pose information and propose an expressive relation representation that captures the relation context, human-object and detailed local parts. We develop a multi-branch deep neural network, referred to as {PMFNet}, to learn such an HOI representation and predict categories of HOI instances. Below we first present an overview of our problem setup and method pipeline in Sec.~\ref{sec:overview}, followed by the detailed description of our model architecture in Sec.~\ref{sec:model}. Finally, Sec.~\ref{sec:train} outlines the model training procedure.

\begin{figure*}[th!]
	\centering
	\includegraphics[width=0.95\linewidth]{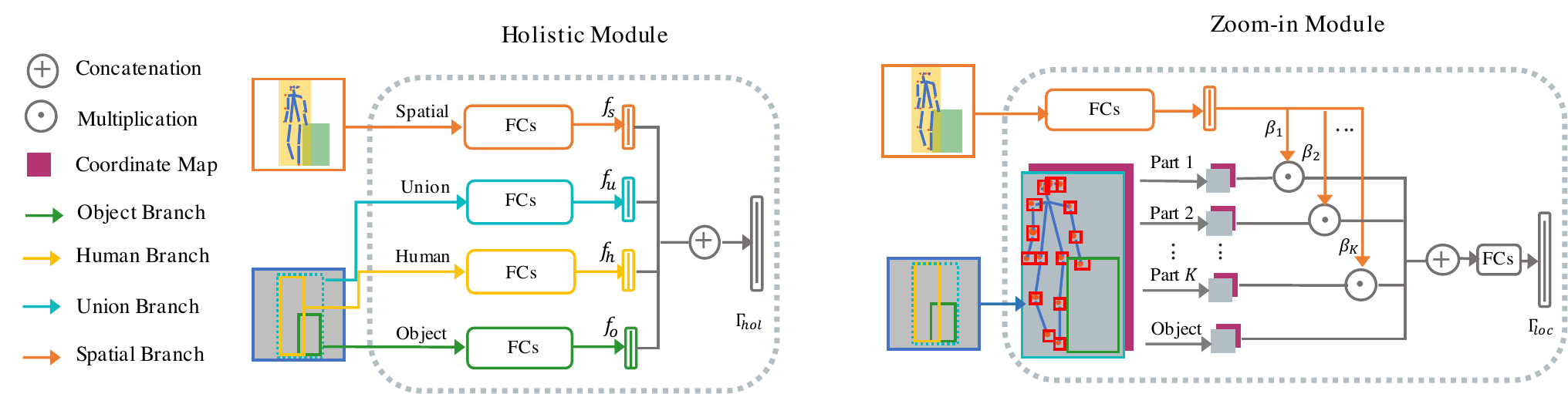}
	\caption[]{The structure of holistic module and zoom-in module. Holistic module includes  human, object, union and spatial branches. Zoom-in module uses human part information and attention mechanism to capture more details.}\vspace{-4mm}
	\label{fig:model1} 
\end{figure*}
\subsection{Overview}\label{sec:overview}
Given an image $ I $, the task of human-object interaction detection aims to generate tuples $\{\langle \mathbf{x}_h, \mathbf{x}_o, c_o, a_{h,o} \rangle\}$ for all HOI instances in the image. Here $\mathbf{x}_h\in \mathbb{R}^4$ denotes human instance location (i.e., bounding box parameters), $\mathbf{x}_o\in \mathbb{R}^4$ denotes object instance location, $c_o \in \{1,...,C\}$ is the object category, and $a_{h,o}\in \{1,...,A\}$ denotes the interaction class associated with $\mathbf{x}_h$ and $\mathbf{x}_o$. For a pair $\langle\mathbf{x}_h, \mathbf{x}_o\rangle$, we use $c_{h,o}^{a} \in \{0,1\} $ to indicate the existence of interaction class $a$.  The object and relation set $\mathcal{C}=\{1,...,C\}$ and $\mathcal{A}=\{1,...,A\}$ are given as inputs to the detection task.  

We adopt a hypothesize-and-classify strategy in which we first generate a set of human-object proposals and then predict their relation classes. In the proposal generation stage, we apply an object detector (e.g., Faster R-CNN~\cite{ren2015faster}) to the input image and obtain a set of human proposals with detection scores $ \{ \langle \mathbf{x}_h, s_h \rangle \} $, and object proposals with their categories and detection scores $ \{ \langle \mathbf{x}_o, c_o, s_o \rangle \} $. Our HOI proposals are generated by pairing up all the human and object proposals. In the relation classification stage, we first estimate a relation score $s_{h,o}^{a}$ for each interaction $a$ and a given $\langle\mathbf{x}_h, \mathbf{x}_o\rangle$ pair. The relation score is then combined with the detection scores of relation entities (human and object) to produce the final HOI score $ R_{h,o}^a $ for the tuple $\{\langle \mathbf{x}_h, \mathbf{x}_o, c_o, a_{h,o} \rangle\}$ as follows,
\begin{align}
R_{h,o}^a= s_{h,o}^a\cdot s_h\cdot s_o,
\end{align}   
where we adopt a soft score fusion by incorporating human score $s_h$ and object score $s_o$ at the same time, which represent detection quality of each proposal.

The main focus of this work is to build a pose-aware relation classifier for predicting the relation score $s_{h,o}^{a}$ given a $\langle\mathbf{x}_h, \mathbf{x}_o\rangle$ pair. To achieve this, we first apply an off-the-shelf pose estimator~\cite{cpn} to a cropped region of proposal $ \mathbf{x}_h $, which generates a pose vector $\mathbf{p}_h = \{p_h^1,...,p_h^K\}$, where $p_h^k \in \mathbb{R}^2$ is $k$-th joint location and $K$ is the number of all joints. In order to incorporate interaction context, human-object and detailed semantic part cues into relation inference, we then introduce a multi-branch deep neural network to generate the relation scores:
\begin{align}
P(c_{h,o}^a=1|I) \propto s_{h, o}^{a} = \mathcal{F}_a(I, \mathbf{x}_h, \mathbf{x}_o, \mathbf{p}_h)
\end{align}
where the network $\mathcal{F}_a$ composes of four modules: a backbone module, a holistic module, a zoom-in module and a fusion module. Below we will describe the details of our model architecture.

\subsection{Model Architecture}\label{sec:model}
Our deep network, PMFNet, instantiates a multi-level relation reasoning with the following four modules: a) a \textit{backbone} module computes image feature map and generates human-object proposals plus spatial configurations; b) a \textit{holistic} module extracts object-level and context features of the proposals; c) a \textit{zoom-in} module focuses on mining part-level features and interaction patterns between human parts and object; and d) a \textit{fusion} module combines both object-level and part-level features to predict the interaction scores. An overview of our model is shown in Fig.~\ref{fig:model}.
\vspace{-2mm}
\subsubsection{Backbone Module}\label{backbone}

We adopt ResNet-50-FPN~\cite{lin2017feature} as our convolutional network to generate feature map $\bm{\Gamma}$ with channel dimension of $ D $. For proposal generation, we use Faster R-CNN~\cite{ren2015faster} as object detector to produce relation proposal pairs $\{\langle \mathbf{x}_h, \mathbf{x}_o\rangle\}$. As mentioned earlier, we also compute human pose vector $\mathbf{p}_h$ for each human proposal $\mathbf{x}_h$ and take it as one of the inputs to our network.

In addition to the conv features, we also extract a set of geometric features to encode the spatial configuration of each human-object instance. We start with two binary masks of human and object proposal in their union space to capture object-level spatial configuration as in~\cite{HOI-DET, gao2018ican}. Moreover, in order to capture fine-level spatial information of human parts and object, we add an additional pose map with predicted poses following~\cite{li2018transferable}. Specifically, we represent the estimated human pose as a line-graph in which all the joints are connected according to the skeleton configuration of COCO dataset. We rasterize the line-graph using a width of $w=3$ pixels and a set of intensity values ranging from 0.05 to 0.95 in a uniform interval to indicate different human parts. Finally, the binary masks and pose map in union space are rescaled to $M\times M$ and concatenated in channel-wise to generate a spatial configuration map.
\vspace{-2mm}
\subsubsection{Holistic Module}\label{holistic}
In order to capture object-level and relation context information, the holistic module is composed of four basic branches: human branch, object branch, union branch and spatial branch, illustrated in Fig.~\ref{fig:model1} (left). The input features of human, object and union branches are cropped from convolution feature map $\bm{\Gamma}$ by applying RoI-Align~\cite{he2017mask} according to human proposal $\mathbf{x}_h$, object proposal $\mathbf{x}_o$ and their union proposal $\mathbf{x}_u$. $\mathbf{x}_u$ is defined as the minimum box in spatial region that contains both $\mathbf{x}_h$ and $\mathbf{x}_o$. Then human features, object features and union features are rescaled to $R_h\times R_h$ resolution. The input of spatial branch directly comes from spatial configuration map generated in Sec.~\ref{backbone}. For each branch, two fully connected layers are adopted to embed the features to an output feature representation. We denote the output features of human, object, union and spatial feature as $f_h$, $f_o$, $f_u,f_s$, and all the features are concatenated to obtain the final holistic feature $\Gamma_{hol}$:
\begin{equation}
	\Gamma_{hol} = f_h\oplus f_o\oplus f_u \oplus f_s
\end{equation}
where $\oplus$ denotes concatenation operation.
\vspace{-2mm}
\subsubsection{Zoom-in Module}\label{sec:zoom}

While the holistic features provide coarse-level information for interactions, many interaction types are defined at a fine-grained level which require detailed local information of human part or object. Hence we design a zoom-in (ZI) module to zoom into human parts to extract part-level features. The overall zoom-in module can be viewed as a network that takes human pose, object proposal and convolution feature map as inputs and extract a set of local interaction features for the HOI relations:

\begin{equation}
	\Gamma_{loc}=\mathcal{F}_{{\rm ZI}}(\mathbf{p}_h, \mathbf{x}_o, \bm{\Gamma})
\end{equation}

Our zoom-in module, illustrated in Fig.~\ref{fig:model1} (right), consists of three components: i) A part-crop component that aims to extract fine-grained human parts features; ii) A {spatial align} component that assigns spatial information to human parts features; iii)
A {semantic attention} component that enhances the human part features relevant to interaction and suppress irrelevant ones.

\vspace{-0.3cm}
\paragraph{Part-crop component}
Given the human pose vector $\mathbf{p}_h = \{p_h^1,...,p_h^K\} $, we define a local region $\mathbf{x}_{p_k}\in \mathbb{R}^4$ for each joint $p_h^k$,  which is a box centered at $p_h^k$ and has a size $\gamma$ proportional to the size of human proposal $\mathbf{x}_h$. Similar to Sec.~\ref{holistic}, we adopt RoI-Align~\cite{he2017mask} for those created part boxes together with object proposal $\mathbf{x}_o$ to generate ($K+1$) regions and rescale to a resolution of $R_p\times R_p$. We denote the pooled part features and object feature as $f_p = \{f_{p_1}, ...,f_{p_K}\}$ and $f_{p_o}$ where
each feature is of size $R_p\times R_p\times D$.

\vspace{-0.3cm}
\paragraph{{Spatial align component}}
Our zoom-in module aims to extract fine-level features of local part regions and model the interaction patterns between human parts and objects. Many interactions have strong correlations with spatial configuration of human parts and object, which can be encoded by the relative locations between different human part and target object. For example, if the target object is close to `hand', the interaction are more likely to be `hold' or `carry', and less likely to be `kick' or `jump'. Based on this observation, we introduce the spatial offset of $ x, y $ coordinates relative to object center as an additional spatial feature for each part. 

In particular, we generate a coordinate map $\bm{\alpha}$ with the same spatial size as the convolution feature map $\bm{\Gamma}$. The map $\bm{\alpha}$ consists of two channels, indicating the $ x $ and $ y $ coordinates for each pixel in $\bm{\Gamma}$, and normalized by the object center. Then we apply RoI-Align~\cite{he2017mask} for each human part $\mathbf{x}_{p_k}$ and object proposal $\mathbf{x}_o$ on $\bm{\alpha}$ and get the spatial map $\alpha_k$ for part $ k $ and $\alpha_o$ for object.
We concatenate the spatial map with the part-crop features so that for a $R_p\times R_p$ cropped part region, we align relative spatial offset to each pixel, which augments part features with a fine-grained spatial cues. The final $k$-th human part feature and object feature are :
\begin{equation}
f'_{p_k} = f_{p_k}\oplus\alpha_k, \quad f'_{p_o} = f_{p_o}\oplus \alpha_o 
\end{equation}
where $f'_{p_k}, f'_{p_o} \in \mathbb{R}^{R_p\times R_p\times (D+2)}$ and $\oplus$ is the concatenate operation.
\vspace{-0.3cm}

\paragraph{Semantic attention component}
As the pose representation also encodes the semantic class of human parts, which typically have strong correlations with interaction types (e.g., `eyes' are important for `read' a book). 
We thus predict a semantic attention using the same spatial configuration map from Sec.~\ref{backbone}.

Our semantic attention network consists of two fully connected layers. A ReLU layer is adopted after the first layer, and a Sigmoid layer is used after the second layer to normalize the final prediction to $[0,1]$. We denote our inferred semantic attention as $ \bm{\beta} \in \mathbb{R}^K $. Note that we do not predict semantic attention for the object, and assume the object has always an attention of value 1, which means it is uniformly important across different instances. The semantic attention is used to weight the part features as follows: 
\begin{equation}
	f''_{p_k} = \beta_k \odot f'_{p_k}
\end{equation}
where $\beta_k\in [0,1]$ is the $ k $-th value of $\bm{\beta}$, $\odot$ indicates element-wise multiplication.

Finally, we concatenate human part features and object feature to obtain the attended part-level features $f_{att} $ and feed it to multiple fully-connected layers ($\mathcal{FC}$) to extract final local feature $\Gamma_{loc}$:
\begin{align}
f_{att} &= f''_{p_1}\oplus \dots f''_{p_K}\oplus f'_{p_o}\\
\Gamma_{loc} &= \mathcal{FC}(f_{att} )
\end{align}


\vspace{-2mm}
\subsubsection{Fusion Module}\label{sec:fuse} 
In order to compute the score $s_{h,o}^{a}$ of pair $\langle\mathbf{x}_h, \mathbf{x}_o \rangle$ for each interaction $a$, we employ a fusion module to fuse relation reasoning from different levels. Our fusion module aims to achieve the following two different goals. First, it uses the coarse-level features as a context cue to determine whether any relation exists for a human-object proposal. This allows us to suppress many background pairs and improve the detection precision. Concretely, we take the holistic feature $\Gamma_{hol}$ and feed it into a network branch consisting of a two-layer fully-connected network followed by a sigmoid function $\sigma$, which generates an  interaction affinity score $s_G$:
\begin{equation}
s_G = \sigma(\mathcal{FC}(\Gamma_{hol})).
\end{equation}
Second, the fusion module use the object-level and part-level features to determine the relation score based on the fine-grained representation. Using a similar network branch, we compute a local relation score $s_L$ from all the relation features:
\begin{equation}
s_L^a = \sigma(\mathcal{FC}_a(\Gamma_{loc}\oplus\Gamma_{hol}))
\end{equation}
where $a$ indicates the relation types. 

Finally, we fuse those two scores defined above to obtain the relation score for a human-object proposal $\langle\mathbf{x}_h, \mathbf{x}_o\rangle$:
\begin{align}
s_{h,o}^a = s_L^a\cdot s_G, \quad \forall a\in \mathcal{A}.
\end{align}  

\subsection{Model Learning}\label{sec:train}
In training stage, we freeze ResNet-50 in our backbone module, and train FPN and other components in Sec.~\ref{sec:model}  in an end-to-end manner. Note that the object detector (Faster R-CNN~\cite{ren2015faster}) and the pose estimator (CPN~\cite{cpn}) are external modules and thus do not participate in learning process.

Assume we have a training set of size $ N $ with relation labels set $Y=\{\textbf{y}^i\}$ and interaction affinity label set $Z=\{z^i\}$, where $\textbf{y}^i = (y^{1,i},...,y^{A,i}) \in \mathds{1}^A$ indicates ground truth relation label for $i$-th sample, and $z^i \in \{0, 1\}$ indicates the relatedness of this sample, $i\in \{1,...,N\}$. We define $z^i = 1$ if $ \exists a \in  \mathcal{A}, y^{a,i} = 1$, else $z^i = 0$.

Suppose our predicted local relation scores are $S_{L}=\{\textbf{s}_{L}^{i}\}$ and affinity scores are $S_{G}=\{  s_{G}^{i} \}$ for those samples, where $\textbf{s}_{L}^{i} = (s_{L}^{1,i},...,s_{L}^{A,i}) $ indicates the predicted local scores of all interactions and $s_{G}^{i} $ is predicted interaction affinity score for $i$-th sample.  As our classification task is actually a multi-label classification problem, we adopt a binary cross entropy loss for each relation class and interaction affinity. Let $L_{CrossEntropy}(a,b)=a\log(b) + (1-a)\log(1-b)$, the overall objective function for our training $L$ is defined as: 
\begin{align}
L = \frac{1}{N}\sum_{i=1}^{N}\big\{&\sum_{a=1}^{A}L_{CrossEntropy}(y^{a, i}, s_{L}^{a, i})\nonumber \\
&+\mu L_{CrossEntropy}(z^{i},s_{G}^{i})\big\} 
\end{align}
where $\mu$ is a hyperparameter to balance the relative importance of multi-label interaction prediction and binary interaction affinity prediction.

\vspace{-2mm}
\section{Experiments}
In this section, we first describe the experimental setting and implementation details. We then evaluate our models with quantitative comparisons to the state-of-the-art approaches, followed by ablation studies to validate the components in our framework. Finally, we show several qualitative results to demonstrate the efficacy of our method.


\subsection{Experimental Setting}
\paragraph{Datasets} We evaluate our method on two HOI benchmarks: V-COCO~\cite{VCOCO} and HICO-DET~\cite{HOI-DET}. V-COCO is a subset of MS-COCO~\cite{MSCOCO}, including 10,346 images (2,533 for training, 2,867 for validation and 4,946 for test) and 16,199 human instances. Each person is annotated with binary labels for 26 action categories. HICO-DET consists of 47,776 images with more than 150K human-object pairs (38,118 images in training set and 9,658 in test set). It has 600 HOI categories over 80 object categories (as in MS-COCO~\cite{MSCOCO}) and 117 unique verbs. 

\vspace{-3mm}
\paragraph{Evaluation Metric} We follow the standard evaluation setting in~\cite{HOI-DET} and use mean average precision to measure the HOI detection performance. We consider an HOI detection as true positive when its predicted bounding boxes of both human and object overlap with the ground-truth bounding boxes with IOUs greater than $0.5$, and the HOI class prediction is correct.

\subsection{Implementation Details}
We use Faster R-CNN~\cite{ren2015faster} as object detector and CPN~\cite{cpn} as pose estimator, which are pre-trained on the COCO train2017 split. 
Each human pose has a total of $K=17$ keypoints as in COCO dataset. 

Our backbone module uses ResNet-50-FPN~\cite{lin2017feature} as feature extractor, and we crop RoI features from the highest resolution feature map in FPN~\cite{lin2017feature}. The size of our spatial configuration map $M$ is set to 64. The RoI-Align in holistic module has a resolution $R_h=7$, while in zoom-in module, the size of human parts is $\gamma=0.1$ of human box height and all the features are rescaled to $R_p=5$.

We freeze ResNet-50 backbone and train the parameters of FPN component. We use SGD optimizer for training with initial learning rate 4e-2, weight decay 1e-4, and momentum 0.9. 
The ratio of positive and negative samples is 1:3. For V-COCO~\cite{VCOCO}, we reduce the learning rate to 4e-3 at iteration 24k, and stop training at iteration 48k.  For HICO-DET~\cite{qi2018learning}, we reduce the learning rate to 4e-3 at iteration 250k and stop training at iteration 300k. During testing, we use object proposals from~\cite{gao2018ican} for fair comparison.  See Suppl. Material for more details.

\subsection{Quantitative Results}
\begin{table}
        \centering
        \resizebox{0.28\textwidth}{!}{
        \begin{tabular}{cc}
        \hline
        Methods            & $AP_{role}$ \\ 
        \hline
        \hline
    
        Gupta \emph{et al.}~\cite{VCOCO}            &31.8 \\
        InteractNet~\cite{gkioxari2018detecting}    &40.0 \\
        GPNN~\cite{qi2018learning}                  &44.0  \\
        iCAN w/late(early)~\cite{gao2018ican}       &44.7 (45.3)\\
        Li \emph{et al.} (${\rm RP_DC_D}$)~\cite{li2018transferable}&47.8\\
        \hline
        Our baseline                            &48.6\\
        Our method (PMFNet)                            &\textbf{52.0}\\
        \hline
        \end{tabular}}
        \vspace{1mm}
        \caption{Performance comparison on V-COCO~\cite{VCOCO} test set.}
        \vspace{-5mm}
        \label{VCOCO}
\end{table}

We compare our proposed framework with several existing approaches for evaluation. We take only  human, object and union branches in holistic module as our baseline, 
while our final model integrates all the modules in Sec.~\ref{sec:model}.

For \textbf{V-COCO} dataset, we evaluate $AP_{role}$ of 24 actions with roles as in~\cite{VCOCO}. As shown in Tab.~\ref{VCOCO}, our baseline method achieves \textbf{48.6} mAP, outperforming all the existing approaches~\cite{VCOCO,gkioxari2018detecting,qi2018learning,gao2018ican,li2018transferable}. Compared to those methods, our baseline adds a union region feature to capture context information, which turns out to be very effective for predicting interaction patterns in a small dataset like V-COCO. 
Moreover, our overall model achieves \textbf{52.0} mAP, which outperforms all the current state-of-the-art methods by a sizable margin, and further improves our baseline by \textbf{3.4} mAP.

For \textbf{HICO-DET}, we choose six current state-of-the-art methods~\cite{li2018transferable,shen2018scaling,HOI-DET,gkioxari2018detecting,qi2018learning,gao2018ican} for comparison. As shown in Tab.~\ref{HICO}, our baseline still performs well and surpasses most existing works except~\cite{li2018transferable}. One potential reason is that HICO-DET dataset has a more fine-grained labeling of interactions (117 categories) than V-COCO (24 categories), and hence the object-level cue is insufficient to distinguish the subtle difference between similar interactions. In contrast, our full model achieves the-state-of-art performance with \textbf{17.46} mAP and \textbf{20.34} mAP on \textbf{Default} and \textbf{Know Object} categories respectively, outperforming all the existing works. In addition, it further improves our baseline by \textbf{2.54} mAP and \textbf{1.51} mAP on \textbf{Default} and \textbf{Know Object} modes respectively. 

Furthermore, 
we divide the 600 HOI categories of the HICO-DET benchmark into two groups as in ~\cite{li2018transferable}: Interactiveness (520 non-trivial HOI classes) and No-interaction (80 no-interaction classes for human and each of 80 object categories). We show the performance of our full model on those two groups compared with our baseline in Tab.~\ref{HICO1}. It is evident that our method achieves larger improvement on the Interactiveness group. As the No-interaction group consists of background classes only, this indicates that our pose-aware dynamic attention is more effective on the challenging task of fine-grained interaction classification.

\begin{table}
	\centering
	\resizebox{0.5\textwidth}{!}{
		\begin{tabular}{ccccccc}
			\hline
			& \multicolumn{3}{c}{Default} & \multicolumn{3}{c}{Know  Object}\\
			Methods            &Full&Rare&Non-Rare    &Full&Rare&Non-Rare  \\  
			\hline
			\hline
			Shen \emph{et al.}~\cite{shen2018scaling} &6.46&4.24&7.12    & - &-&-     \\
			HO-RCNN~\cite{HOI-DET}                     &7.81&5.37&8.54 &10.41&8.94&10.85 \\
			InteractNet~\cite{gkioxari2018detecting} &9.94&7.16&10.77   &-&-&-    \\
			GPNN~\cite{qi2018learning}         &13.11&9.34&14.23  &-&-&-   \\
			iCAN~\cite{gao2018ican}            &14.84&10.45&16.15 &16.26&11.33&17.73   \\
			Li \emph{et al.}-${\rm RP_DC_D}$~\cite{li2018transferable}   &17.03&13.42&\textbf{18.11} &19.17&15.51&20.26   \\
			\hline
			Our baseline                      &14.92&11.42&15.96   &18.83&15.30&19.89\\
			Our method (PMFNet)                      &\textbf{17.46}&\textbf{15.65}&18.00   &\textbf{20.34}&\textbf{17.47}&\textbf{21.20}\\
			\hline
		\end{tabular}}
		\vspace{0.01mm}
		\caption{Results Comparison on HICO-DET~\cite{HOI-DET} test set.}
		\vspace{-1mm}
		\label{HICO}
	\end{table}
	
	\begin{table}
		\centering
		\resizebox{0.42\textwidth}{!}{
			\begin{tabular}{ccc}
				\hline
				& \multicolumn{2}{c}{Default} \\
				Methods        &Interactiveness(520) &No-interaction(80)  \\  
				\hline
				
				Our baseline  & 15.97 & 8.05  \\
				Our method (PMFNet)   &18.79 & 8.83  \\
				\hline
			\end{tabular}}
			\vspace{1mm}
			\caption{Improvements of our model in Interactiveness and No-interaction HOIs on HICO-DET~\cite{HOI-DET} test set.} 
			\vspace{-5mm}
			\label{HICO1}
		\end{table}

\begin{figure*}[ht]
    \centering
    \includegraphics[width=1.0\linewidth]{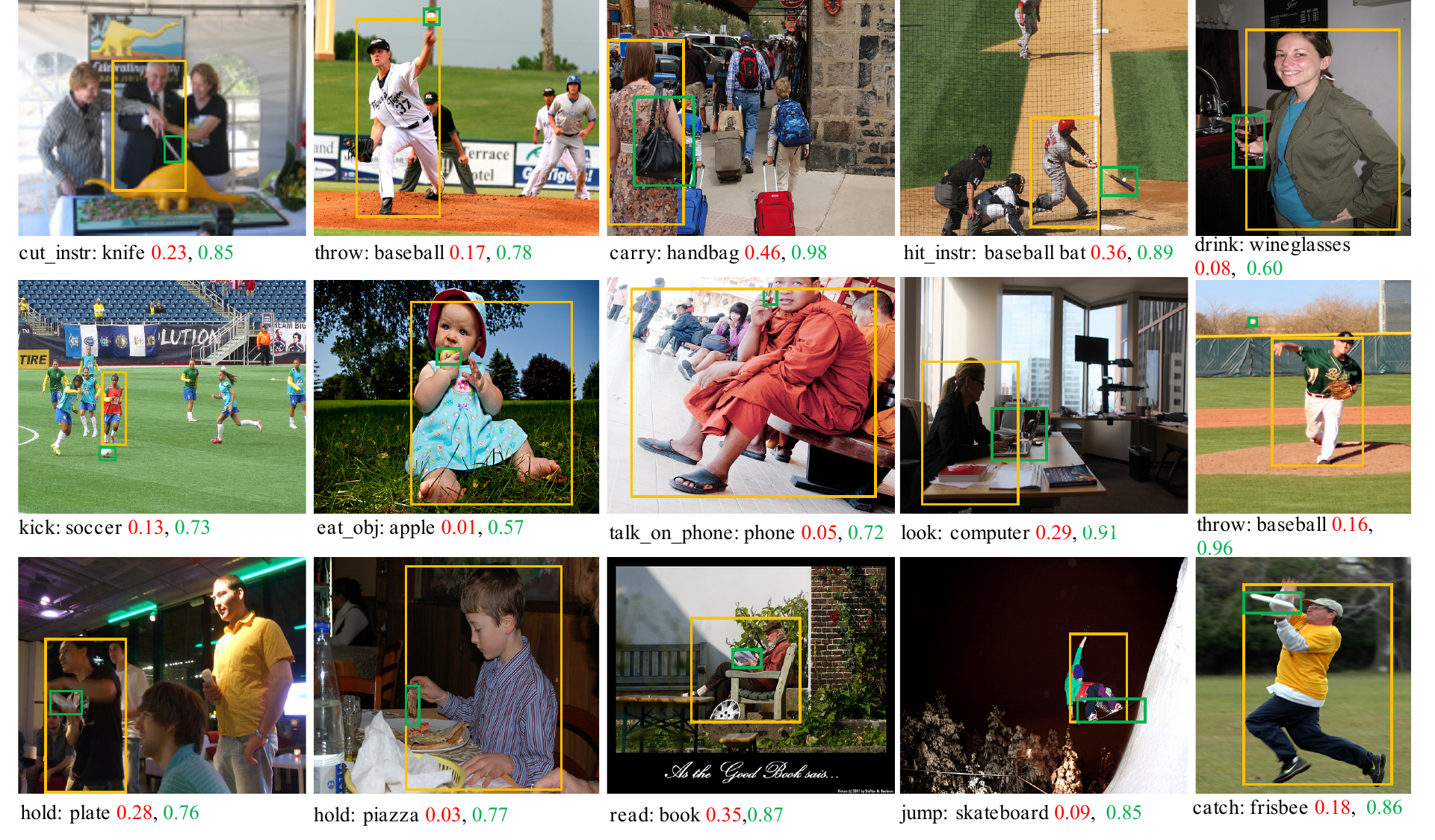}
    \caption{HOI detection results compared with baseline on V-COCO\cite{VCOCO} val set. For each ground-truth interaction, we compare interaction action score with baseline method. {\color{red}{Red number}} and {\textcolor[RGB]{0,176,80}{green number}} denote score predicted by baseline and our approach respectively. As shown in figure, our approach can be more confident for predicting interaction actions when the target objects are very small and ambiguous (all score improvements great than 0.5). }\vspace{-3mm}
    \label{comp_score}
\end{figure*}

\begin{figure}[ht]
    \centering
    \includegraphics[width=0.9\linewidth]{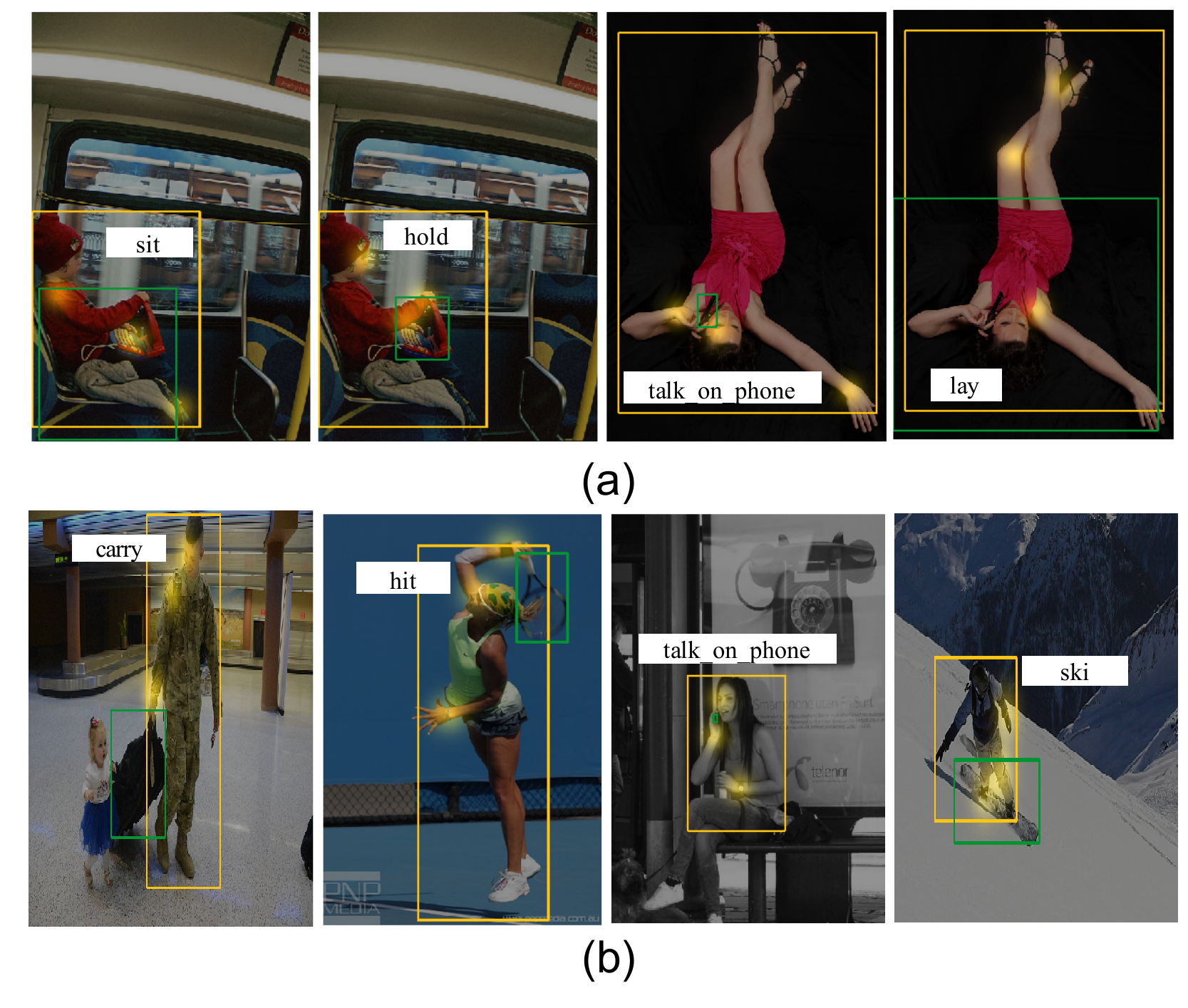}
    \caption{Semantic Attention on (a) the same person interacting with different objects, and (b) different person with various interactions.}\vspace{-5mm}
    \label{SA1}
\end{figure}

\subsection{Ablation Study}
In this section, we perform several experiments to evaluate the effectiveness of our model components on V-COCO dataset (Tab.~\ref{V-COCO-Aba}).

\vspace{-3mm}
\paragraph{Spatial Configuration Map (SCM)}  As in~\cite{li2018transferable}, we augment human and object binary masks with an additional human pose configuration map, which provides more detailed spatial information on human parts. This enriched SCM enables the network to filter out non-interactive human-object instances more effectively. As shown in Tab.~\ref{V-COCO-Aba}, the SCM improves our baseline by \textbf{0.7} mAP.


\vspace{-3mm}
\paragraph{Part-crop (PC)} Part-crop component zooms into semantic human parts and provides fine-grained feature representation of human body. Experiments in Tab.~\ref{V-COCO-Aba} show the effectiveness of zoom-in parts feature, which improves mAP from \textbf{49.9} to \textbf{51.0}. We note that the following spatial align and semantic attention component are built on top of the part-crop component.


\begin{table}
	\centering
	\resizebox{0.45\textwidth}{!}{
		\begin{tabular}{ccccccc}
			\hline
			& \multicolumn{5}{c}{Components}&   \\
			Methods     & SCM&PC&SpAlign&SeAtten&IA     & $AP_{role}$  \\ 
			\hline
			Baseline    &-&-&-&-&-                     &49.2  \\
			\hline
			\multirow{5}{*}{Incremental} &\checkmark&-&-&-&-            &49.9  \\
			&\checkmark&\checkmark&-&-&-                      &51.0  \\
			&\checkmark&\checkmark&\checkmark&-&-                  &52.4  \\
			&\checkmark&\checkmark&\checkmark&\checkmark&-&52.7  \\
			\hline
			\multirow{5}{*}{Drop-one-out}	&-&\checkmark&\checkmark&\checkmark&\checkmark         &52.0  \\
			&\checkmark&-&-&-&\checkmark       &50.3  \\
			&\checkmark&\checkmark&-&\checkmark&\checkmark                 & 51.1 \\
			&\checkmark&\checkmark&\checkmark&-&\checkmark&52.6  \\
			&\checkmark&\checkmark&\checkmark&\checkmark&-& 52.7  \\
			\hline
			\vspace{0.mm}
			Our method (PMFNet) &\checkmark&\checkmark&\checkmark&\checkmark&\checkmark& \textbf{53.0}  \\
			\hline
	\end{tabular}}
	\vspace{1mm}
	\caption{Ablation study on V-COCO~\cite{VCOCO} val set.} 
	\vspace{-5mm}
	\label{V-COCO-Aba}
\end{table}

\vspace{-3mm}
\paragraph{Spatial Align (SpAlign)} Spatial align component computes relative locations of all the parts w.r.t. the object and integrates them into the part features, which captures a `normalized' local context. We observe a significant improvement from \textbf{51.0} to \textbf{52.4} in Tab.~\ref{V-COCO-Aba}.

\vspace{-3mm}
\paragraph{Semantic Attention (SeAtten)} The semantic attention focuses on informative human parts and suppress other irrelevant ones. Its part-wise attention scores provides an interpretable feature for our predictions. As shown in Tab.~\ref{V-COCO-Aba}, SeAtten slightly improves the preformance by \textbf{0.3} mAP.

\vspace{-3mm}
\paragraph{Interaction Affinity (IA)} Similar to~\cite{li2018transferable}, the interaction affinity indicates whether a human-object pair have interaction, and can reduce false positives by lowering their interaction scores. We can observe from Tab.~\ref{V-COCO-Aba} that IA improves performance by \textbf{0.3} mAP.



\vspace{-3mm}
\paragraph{Drop-one-out Ablation study} We further perform a drop-one-out ablation study which all independent components are removed individually, shown in tab.~\ref{V-COCO-Aba}. The results demonstrate that each independent component indeed contribute to our final performance.

\subsection{Qualitative Visualization Results}
Fig.~\ref{comp_score} shows our HOIs detection results compared with the baseline approach. We can see that our framework is capable of detecting difficult HOIs where the target objects are very small and generates a more confident score. This suggests that part-level features provide more informative visual cues for difficult human-object interaction pairs.

Fig.~\ref{SA1} visualizes semantic attention on a variety of HOI cases, each of which provides an interpretable outcome for our predictions. The highlighted joint regions indicate that our Semantic Attention (SeAtten) component generates an attention score higher than 0.7 for the related keypoint. 
In Fig.~\ref{SA1}(a), for the same person interacting with various target objects, our SeAtten component is capable of automatically focusing on different human parts that are strongly related to interaction action. As the two images in up-left show, when the child interacting with chair, SeAtten will concentrate on the full body joints; while he interacting with an instrument, SeAtten will focus on his hands. To validate the generalization capacity of the SeAtten component, we also visualize several other HOI examples in Fig.\ref{SA1}(b). For different persons with various interactions, our SeAtten component can always produce  meaningful highlight on human parts which are relevant to each interaction type.


\vspace{-2mm}
\section{Conclusion}

In this paper, we have developed an effective multi-level reasoning approach to human-object interaction detection. Our method is capable of incorporating interaction level, visual object level and human parts level features under the guidance of human pose information. As a result, it is able to recognize visual relations with subtle differences. We present a multi-branch deep neural network to instantiate our core idea of multi-level reasoning. Moreover, we introduce a semantic part-based attention mechanism at the part level to automatically extract relevant human parts for each interaction instance. The visualization of our attention map produces an interpretable output for the human-object relation detection task. Finally, we achieve the state-of-the-art performances on both V-COCO and HICO-DET benchmarks, and outperform other approaches by a large margin on V-COCO dataset.

\vspace{-2mm}
\section*{Acknowledgement}
This work was supported by Shanghai NSF Grant (No. 18ZR1425100) and NSFC Grant (No. 61703195).

{\small
	\bibliographystyle{ieee_fullname}
	\bibliography{egbib}
}
\end{document}